# Novel Grey Interval Weight Determining and Hybrid Grey Interval Relation Method in Multiple Attribute Decision-Making*

Gol Kim [a]

[a] *Center of Natural Science, University of Sciences, Pyongyang, DPR Korea*
*E-mail:golkim124@yahoo.com*

25 June 2012

## Abstract

This paper proposes a grey interval relation TOPSIS for the decision making in which all of the attribute weights and attribute values are given by the interval grey numbers. The feature of our method different from other grey relation decision-making is that all of the subjective and objective weights are obtained by interval grey number and that decision-making is performed based on the relative approach degree of grey TOPSIS, the relative approach degree of grey incidence and the relative membership degree of grey incidence using 2-dimensional Euclidean distance. The weighted Borda method is used for combining the results of three methods. An example shows the applicability of the proposed approach.

**Keywords:** Grey interval weight, Multiple attribute decision making, Grey interval relation TOPSIS

## 1.Introduction

The multiple attribute decision-making (MADM) problems are of the most interesting problems for many decision-making experts. This problem arises in various fields of the real life, and constitutes very important content in scientific research such as management science, decision-making theory, system theory, operational research and economics.

Now, many effective methods to determine the attributive weights have been studied for MADM. Those are the subjective weight determining methods such as the feature vector method ( Saaty T.L. 1977 ), the least square sum method (Chu A Tw, Kalaba R E, Spingarn K, 1979), Delphi and AHP method (Hwang C.L., Lin M, 1987), and the objective weight determining methods such as the entropy method (Hwang C.L., Yoon K, 1981), the principal component analysis (Yan Jian-huo, 1989 ) and DEA (Data Envelopment Analysis) (Ye Chen, Kevin W. Li, Haiyan Xu and Sifeng Liu, 2009).

The final ranking method affects greatly on the decision-making process. Hwang and Yoon (1981) proposed a new approach, TOPSIS (Technique for Order Preference by Similarity to Ideal Solution) for solving MADM problem. Recently, TOPSIS methods with interval weights (Gao Feng-ji, et al, 2005) and multiple attribute interval number TOPSIS (Chu A Tw, Kalaba R E, Spingarn K, 1979) have been studied. Guo Kai-hong and Mu You-jing (2012) studied the relation between several possibility degree formulas and proposed a possibility degree matrices-based method that aimed to objectively determine the weights of criteria in MADM with intervals. A hybrid approach integrating OWA (Ordered Weighted Averaging) aggregation into TOPSIS is proposed to tackle multiple criteria decision analysis (MCDA) problems (Ye Chen, Kevin W. Li, Si-feng Liu, 2011). A hybrid approach of DEA (Data Envelopment Analysis) and TOPSIS is proposed for MCDA in emergency management (Ye Chen, Kevin W. Li, Haiyan Xu and Sifeng Liu, 2009).

The degree of grey incidence in grey system theory is a very important technical conception. The computational formulas of incidence degree such as grey incidence degree, grey absolute incidence degree and grey comprehensive incidence degree are introduced and the concepts of grey relation decision-making are given (Liu Si-feng , Lin Yi, 2004). Luo Dang, Liu Si-feng et al (2005) extended the traditional grey relation decision-making method to interval grey number, proposed a choosing method of plan based on maximal degree and constructed a formula of grey interval incidence degree and a grey interval relative incidence degree. The ideal optimal plan for MADM problem was defined and a formula of grey interval incidence coefficient was obtained (Dang Yao-Guo, Liu Si-feng et al, 2004). Other methods in grey decision-making are the grey clustering decision-making (Mi Chuan-min, Liu Si-feng et al, 2006) and the grey incidence projection method (Zhang-Chao, et al, 2007).

In the case that the attribute weight information is known partially, decision-maker has to solve the MADM problem with preference to plans. To our knowledge, for the case that both of attribute vector and weight vector is given by interval grey numbers, any method of obtaining all weights by interval numbers has not been studied yet.

This paper considers a hybrid MADM problem with interval attribute and interval decision matrix, and presents a new grey interval relation method which considers comprehensive weight and preference of decision-making. In this paper, first of all, the subjective weights of attributes are obtained as interval number based on group AHP method. Next, in the case which the attribute values are given by interval grey numbers, the objective weights are determined based on optimization method. Besides, in the case which the attribute values are given by interval grey numbers, the objective weights are obtained by interval number based on entropy method. Thus, the comprehensive weights of attribute for decision-making are determined by combining the subjective weight and the objective weight using multiplicative composition method. Therefore, the attribute weights can reflect the subjective and objective information of the system more sufficiently. In this paper, secondly, when all of the attribute weights and attribute values are given by interval grey number, three grey relation decision-making methods are studied such as the evaluation of plan by the relative approach degree of grey TOPSIS, the evaluation of plan by the relative approach degree of grey incidence degree method, and the evaluation of plan by the relative membership degree of grey incidence degree method. The final rank based on rank vectors for each method is obtained by weighted Borda method. Finally, an example is given to show the performance of our method.

## 2. Some concepts and normalization of decision matrix

**[Definition 1]** Assume that $a(\otimes) \in [\underline{a}, \overline{a}]$ and $b(\otimes) \in [\underline{b}, \overline{b}]$ are two interval grey numbers. Then, distance between $a(\otimes) \in [\underline{a}, \overline{a}]$ and $b(\otimes) \in [\underline{b}, \overline{b}]$ is defined by

$$d(a(\otimes), b(\otimes)) = \sqrt{(\overline{b} - \overline{a})^2 + (\underline{b} - \underline{a})^2}.$$

Assume that $A = \{A_1, A_2, \ldots, A_n\}$ is a set of the decision plans and $G = \{G_1, G_2, \ldots, G_m\}$ is an attribute set. The value of the attribute $G_j$ for plan $A_i$ is given the non-negative interval number by $a_{ij}(\otimes) \in [\underline{a}_{ij}, \overline{a}_{ij}]$, $(0 \leq \underline{a}_{ij}, \leq \overline{a}_{ij}, i = \overline{1, n}; j = \overline{1, m})$.

Let $a_i(\otimes) = (a_{i1}(\otimes), a_{i2}(\otimes), \cdots, a_{im}(\otimes)), i = 1, \cdots, n$ be attribute vector and $R(\otimes) = (a_{ij}(\otimes))_{n \times m}$ be decision matrix.

We make normalization processing for $a_{ij}(\otimes)$s as follows.

For the attribute of cost type,

$$\underline{x}_{ij}(\otimes) = \frac{1/\overline{a}_{ij}(\otimes)}{\sum_{i=1}^{n}(1/\underline{a}_{ij}(\otimes))}, \quad \overline{x}_{ij}(\otimes) = \frac{1/\underline{a}_{ij}(\otimes)}{\sum_{i=1}^{n}(1/\overline{a}_{ij}(\otimes))} \quad (i = \overline{1,n}, j = \overline{1,m})$$

For the attribute of effect type,
$$\underline{x}_{ij}(\otimes) = \frac{\underline{a}_{ij}(\otimes)}{\sum_{i=1}^{n}\overline{a}_{ij}(\otimes)}, \quad \overline{x}_{ij}(\otimes) = \frac{\overline{a}_{ij}(\otimes)}{\sum_{i=1}^{n}\underline{a}_{ij}(\otimes)} \quad (i = 1,\cdots,n, j = 1,\cdots,m)$$

**[Definition 2]** Assume that $X = (x_{ij}(\otimes))_{n \times m}$ is normalized decision matrix. The attribute vector of each plan is $x_i(\otimes) = (x_{i1}(\otimes), x_{i2}(\otimes), \cdots, x_{im}(\otimes))$, $i = 1, \cdots, n$, where $x_{ij}(\otimes) \in [\underline{x}_{ij}, \overline{x}_{ij}]$ is non-negative interval grey number on $[0,1]$.

## 3. Determining of attribute weights
### 3.1. Subjective weight determining of attributes

Let $\alpha_l = [\alpha_l^1, \cdots, \alpha_l^j, \cdots, \alpha_l^m]$, $(l = \overline{1,L})$ be the attribute weights determined by AHP from the decision-making group. The weight of attribute $G_j$ is given as interval grey number $\alpha_j(\otimes) \in [\underline{\alpha}_j, \overline{\alpha}_j]$, $0 \leq \underline{\alpha}_j \leq \overline{\alpha}_j$, where

$$\underline{\alpha}_j = \min_{1 \leq l \leq L}\{\alpha_l^j\}, \quad \overline{\alpha}_j = \max_{1 \leq l \leq L}\{\alpha_l^j\}, \quad j = \overline{1,m}.$$

### 3.2. Objective weight of attributes
#### 3.2.1. Objective weight determining by optimization when attributes values are given by interval grey number

We define the deviation of decision plan $A_i$ from all other decision plans for attribute $G_j$ in normalized decision matrix $X = (x_{ij}(\otimes))_{n \times m}$ as follows

$$D_{ij}(\beta^{opt}) = \sum_{k=1}^{n} d(x_{ij}, x_{kj})\beta_j^{opt} = \sum_{k=1}^{n} \sqrt{(\overline{x}_{kj} - \overline{x}_{ij})^2 + (\underline{x}_{kj} - \underline{x}_{ij})^2}\,\beta_j^{opt}$$

In order to choose weight vector $\beta^{opt}$ such that sum of overall deviation for the decision plan attains maximum, we define a deviation function such as

$$D(\beta) = \sum_{j=1}^{m}\sum_{i=1}^{n}\sum_{k=1}^{n} d(x_{ij}(\otimes), x_{kj}(\otimes))\beta_j$$

and solve the following nonlinear programming problem.

[P1] $\quad \max D(\beta) = \sum_{j=1}^{m}\sum_{i=1}^{n}\sum_{k=1}^{n} d(x_{ij}(\otimes), x_{kj}(\otimes))\beta_j$,

$$s.t. \sum_{j=1}^{m}\beta_j^2 = 1, \; \beta_j \geq 0, \; j = 1,\cdots,m$$

**[Theorem 1]** The solution of problem P1 is given by

$$\overline{\beta}_j = \frac{\sum_{i=1}^{n}\sum_{k=1}^{n} d(x_{ij}(\otimes), x_{kj}(\otimes))}{\sqrt{\sum_{j=1}^{m}\left[\sum_{i=1}^{n}\sum_{k=1}^{n} d(x_{ij}(\otimes), x_{kj}(\otimes))\right]^2}}, \; j = 1,\cdots,m. \quad \square$$

By the normalization of $\overline{\beta}_j$, $j = 1, \cdots, m$, we obtain

$$\beta_j^{opt} = \frac{\sum_{i=1}^{n}\sum_{k=1}^{n} d(x_{ij}(\otimes), x_{kj}(\otimes))}{\sum_{j=1}^{m}\sum_{i=1}^{n}\sum_{k=1}^{n} d(x_{ij}(\otimes), x_{kj}(\otimes))}, \quad j = 1, \cdots, m.$$

### 3.2.2. Objective weight determining by entropy method in the case which the attribute values are the interval grey number

The entropy weights of the normalized decision matrix $X = (x_{ij}(\otimes))_{n \times m}$, $x_{ij}(\otimes) \in [\underline{x}_{ij}, \overline{x}_{ij}]$ for lower bound $\underline{x}_{ij}$ and upper bound $\overline{x}_{ij}$ of grey number $x_{ij}(\otimes)$ have to be found, respectively.

First, let's find the entropy weight for lower bound $\underline{x}_{ij}$. Letting

$$\underline{p}_{ij} = \frac{\underline{x}_{ij}}{\sum_{i=1}^{n} \underline{x}_{ij}}, \quad (i = \overline{1,n}, j = \overline{1,m}),$$

the entropy value of $j$ th attribute is given by $\underline{E}_j = -k \sum_{i=1}^{n} \underline{p}_{ij} \ln \underline{p}_{ij}$, $j = \overline{1,m}$, where $k = \dfrac{1}{\ln n}$. In the above formula, if $\underline{p}_{ij} = 0$, then we regard that $\underline{p}_{ij} \ln \underline{p}_{ij} = 0$. Then $0 \le \underline{E}_j \le 1$, $j = 1, \cdots, m$ and the deviation coefficient for $j$ th attribute is given by

$$\underline{\eta}_j = 1 - \underline{E}_j, \quad j = 1, \cdots, m.$$

The entropy weight $\underline{\beta}^{ent} = (\underline{\beta}_1^{ent}, \cdots, \underline{\beta}_j^{ent}, \cdots, \underline{\beta}_m^{ent})$ for lower bound $\underline{x}_{ij}$ is given by

$$\underline{\beta}_j^{ent} = \frac{\underline{\eta}_j}{\sum_{j=1}^{m} \underline{\eta}_j} = \frac{1 - \underline{E}_j}{\sum_{j=1}^{m}(1 - \underline{E}_j)} = \frac{1 - \underline{E}_j}{m - \sum_{j=1}^{m} \underline{E}_j}, \quad j = \overline{1,m}$$

Similarly, the entropy weight $\overline{\beta}^{ent} = (\overline{\beta}_1^{ent}, \cdots, \overline{\beta}_j^{ent}, \cdots, \overline{\beta}_m^{ent})$ for upper bound $\overline{x}_{ij}$ is

$$\overline{\beta}_j^{ent} = \frac{1 - \overline{E}_j}{m - \sum_{j=1}^{m} \overline{E}_j}, \quad j = \overline{1,m},$$

where $\overline{E}_j = -k \sum_{i=1}^{n} \overline{p}_{ij} \ln \overline{p}_{ij}$ and $\overline{p}_{ij} = \dfrac{\overline{x}_{ij}}{\sum_{i=1}^{n} \overline{x}_{ij}}$ $(i = \overline{1,n}, j = \overline{1,m})$.

### 3.2.3. Determining of comprehensive objective weights

The comprehensive objective weight is determined by the interval grey number

$$\beta(\otimes) = (\beta_1(\otimes), \beta_2(\otimes), \cdots, \beta_m(\otimes)), \beta_j(\otimes) \in [\underline{\beta}_j, \overline{\beta}_j],$$

$$\underline{\beta}_j(\otimes) = \min\{\beta_j^{opt}, \underline{\beta}_j^{ent}, \overline{\beta}_j^{ent}\}, \quad \overline{\beta}_j(\otimes) = \max\{\beta_j^{opt}, \underline{\beta}_j^{ent}, \overline{\beta}_j^{ent}\}.$$

### 3.3. Determining of final comprehensive weights

The final comprehensive weight is determined by

$$w_j(\otimes) = \frac{\alpha_j(\otimes) \times \beta_j(\otimes)}{\sum_{j=1}^{m} \alpha_j(\otimes) \times \beta_j(\otimes)}, \quad j = \overline{1,m}$$

where $\alpha_j(\otimes)$ and $\beta_j(\otimes)$ are the subjective weight and the objective weight for $j$ th attribute, respectively. Thus, the weight of the attribute $G_j$ is given by the interval grey number $w_j(\otimes)$ such as $w_j(\otimes) \in [\underline{w}_j, \overline{w}_j], 0 \leq \underline{w}_j \leq \overline{w}_j \leq 1, j = \overline{1,m}$.

## 4. Some evaluation methods of the decision plans

### 4.1. Evaluation of plan by the relative approach degree of grey TOPSIS method

Assume that the subjective preference value of the plan $A_i$ is given by the interval grey number $q_i(\otimes)$, where $q_i(\otimes) \in [\underline{q}_i, \overline{q}_i], 0 \leq \underline{q}_i \leq \overline{q}_i \leq 1, i = \overline{1,n}$.

The normalized decision matrix with the subjective preference is $\tilde{Z} = (z_{ij}(\otimes))_{n \times m}$,

$$z_{ij}(\otimes) = \frac{1}{2}q_i(\otimes) + \frac{1}{2}x_{ij}(\otimes) \in \left[\frac{1}{2}\underline{q}_i + \frac{1}{2}\underline{x}_{ij}, \frac{1}{2}\overline{q}_i + \frac{1}{2}\overline{x}_{ij}\right].$$

Let $\tilde{Y} = (y_{ij}(\otimes))_{n \times m}$ be the comprehensive weighted decision matrix such as

$$y_{ij}(\otimes) = w_j(\otimes) z_{ij}(\otimes) \in [\underline{y}_{ij}, \overline{y}_{ij}], i = \overline{1,n}, j = \overline{1,m}.$$

The attribute vector of each plan for the normalized comprehensive weighted decision matrix is $y_i(\otimes) = (y_{i1}(\otimes), y_{i2}(\otimes), \cdots, y_{im}(\otimes)), i = \overline{1,n}$.

**[Definition 3]** We put

$$\underline{y}_j^+ = \max_{1 \leq i \leq n}\{\underline{y}_{ij}\}, \overline{y}_j^+ = \max_{1 \leq i \leq n}\{\overline{y}_{ij}\}, \underline{y}_j^- = \min_{1 \leq i \leq n}\{\underline{y}_{ij}\}, \overline{y}_j^- = \min_{1 \leq i \leq n}\{\overline{y}_{ij}\}, j = \overline{1,m}.$$

Then, the $m$-dimension interval grey number vector $y^+(\otimes)$ ($y^-(\otimes)$) such as

$$y^+(\otimes) = (y_1^+(\otimes), y_2^+(\otimes), \cdots, y_j^+(\otimes), \cdots, y_m^+(\otimes))$$
$$(y^-(\otimes) = (y_1^-(\otimes), y_2^-(\otimes), \cdots, y_j^-(\otimes), \cdots, y_m^-(\otimes)))$$

is called a positive (negative) ideal plan attribute vector, where $y_j^+(\otimes) \in [\underline{y}_j^+, \overline{y}_j^+]$ ($y_j^-(\otimes) \in [\underline{y}_j^-, \overline{y}_j^-]$), $j = \overline{1,m}$.

Euclidian distance between each plan attribute vector $y_i(\otimes)$ and the positive or negative ideal plan attribute vector $y^+(\otimes)$ or $y^-(\otimes)$ is

$$D_i^+ = \sqrt{\sum_{j=1}^m \left[(\overline{y}_{ij} - \overline{y}_j^+)^2 + (\underline{y}_{ij} - \underline{y}_j^+)^2\right]}$$

or

$$D_i^- = \sqrt{\sum_{j=1}^m \left[(\overline{y}_{ij} - \overline{y}_j^-)^2 + (\underline{y}_{ij} - \underline{y}_j^-)^2\right]}.$$

The relative approach degree between each evaluation plan and the optimal plan is

$$C_i = \frac{D_i^-}{D_i^+ + D_i^-}, i = \overline{1,n}.$$

The best plan is one corresponding to the largest $C_i$.

### 4.2 Evaluation of plan by the relative approach degree of grey incidence

**[Definition 4]** Let $\{y_{ij}(\otimes)\}_{n \times m}$ be the normalized comprehensive weighted decision matrix and $y_j^+(\otimes)$ and $y_j^-(\otimes)$ be the positive and negative ideal plan attribute vector, respectively. We define

$$r_{ij}^{+} = \frac{\min_i \min_j d(y_{ij}(\otimes), y_j^{+}(\otimes)) + \rho \max_i \max_j d(y_{ij}(\otimes), y_j^{+}(\otimes))}{d(y_{ij}(\otimes), y_j^{+}(\otimes)) + \rho \max_i \max_j d(y_{ij}(\otimes), y_j^{+}(\otimes))},$$

$$r_{ij}^{-} = \frac{\min_i \min_j d(y_{ij}(\otimes), y_j^{-}(\otimes)) + \rho \max_i \max_j d(y_{ij}(\otimes), y_j^{-}(\otimes))}{d(y_{ij}(\otimes), y_j^{-}(\otimes)) + \rho \max_i \max_j d(y_{ij}(\otimes), y_j^{-}(\otimes))}.$$

Then, $r_{ij}^{+}$ ($r_{ij}^{-}$) is called the coefficient of positive (negative) ideal grey interval incidence of $y_{ij}(\otimes)$ with respect to the positive ideal attribute value $y_j^{+}(\otimes)$ ($y_j^{-}(\otimes)$), where $\rho \in (0,1)$ and generally $\rho = 0.5$ is taken.

**[Definition 5]** The matrix $P^{+} = \{r_{ij}^{+}\}_{n \times m}$ ($P^{-} = \{r_{ij}^{-}\}_{n \times m}$) is called a grey incidence coefficient matrix of the given plan with respect to the positive (negative) ideal plan.

**[Definition 6]** Let

$$G(y^{+}(\otimes), y_i(\otimes)) = \frac{1}{m}\sum_{j=1}^{m} r_{ij}^{+}, \; G(y^{-}(\otimes), y_i(\otimes)) = \frac{1}{m}\sum_{j=1}^{m} r_{ij}^{-}, i = 1, \cdots, n.$$

Then $G(y^{+}(\otimes), y_i(\otimes))$ ($G(y^{-}(\otimes), y_i(\otimes))$) is called a degree of grey interval incidence of the comprehensive attribute vector for the plan $A_i$ with respect to the positive (negative) ideal plan attribute vector.

**[Theorem 2]** The grey interval incidence degrees $G(y^{+}(\otimes), y_i(\otimes))$ and $G(y^{-}(\otimes), y_i(\otimes))$ satisfy the four axioms of grey incidence degree (Sifeng Liu and Lin Y., 2004), i.e. normality, pair-symmetry, wholeness and closeness.

We define a degree of grey incidence relative approach by

$$C_i = \frac{G(y^{+}(\otimes), y_i(\otimes))}{G(y^{+}(\otimes), y_i(\otimes)) + G(y^{-}(\otimes), y_i(\otimes))} \; (i = \overline{1,n})$$

Then, $0 < G(y^{+}(\otimes), y_i(\otimes)) \leq 1$, $0 < G(y^{-}(\otimes), y_i(\otimes)) \leq 1$ and $0 < C_i < 1$.

The degree of grey incidence relative approach is modified by introducing the preference coefficients as follows.

$$C_i = \begin{cases} \dfrac{G(y^{+}(\otimes), y_i(\otimes)) \cdot \theta_{+}}{G(y^{+}(\otimes), y_i(\otimes)) \cdot \theta_{+} + G(y^{-}(\otimes), y_i(\otimes)) \cdot \theta_{-}} ; & 0 < \theta_{+} < 1, \theta_{-} < 1 \\ G(y^{+}(\otimes), y_i(\otimes)) ; & \theta_{+} = 1, \theta_{-} = 0 \end{cases},$$

where $\theta_{+}$ and $\theta_{-}$ are the preference coefficients, respectively. Generally, we regard as $\theta_{+} > \theta_{-}$ and choose it so as to satisfy $0 < \theta_{+} \leq 1$, $0 < \theta_{-} \leq 1$, $\theta_{+} + \theta_{-} = 1$. If $\theta_{+} + \theta_{-} \neq 1$, then we normalize those by

$$\theta_{+}' = \frac{\theta_{+}}{\theta_{+} + \theta_{-}}, \theta_{-}' = \frac{\theta_{-}}{\theta_{+} + \theta_{-}}.$$

to obtain $\theta_{+} + \theta_{-} = 1$. When $\theta_{+} = \theta_{-} = \frac{1}{2}$, it becomes the canonical formula for the degree of grey incidence relative approach.

The optimal plan corresponds to the largest value among of the relative approach degree $C_i$.

### 4.3. Evaluation of plan by the relative membership degree of grey incidence

If the membership degree of the positive ideal plan with respect to the plan $A_i$ is $u_i$, the membership degree of the negative ideal plan corresponding to the plan $A_i$ is $1 - u_i$.

Therefore, we can find the membership degree vector $u = (u_1, u_2, \cdots, u_n)$ by solving the following problem.

[P2] $\quad \min F(u) = \sum_{i=1}^{n} \left\{ [(1-u_i)G(y^+(\otimes), y_i(\otimes))]^2 + [u_i G(y^-(\otimes), y_i(\otimes))]^2 \right\}$.

**[Theorem 3]** The optimal solution of the optimization problem P2 is given by

$$u_i = \frac{G^2(y^+(\otimes), y_i(\otimes))}{G^2(y^+(\otimes), y_i(\otimes)) + G^2(y^-(\otimes), y_i(\otimes))} \quad (i = \overline{1,n})$$

The optimal plan is one corresponding to the largest membership degree $u_i$.

In practice, the weights of the incidence coefficient are not always equal. Therefore, we find the weighted incidence degrees such as

$$G(y^+(\otimes), y_i(\otimes)) = \sum_{j=1}^{m} \gamma_j r_{ij}^+, \quad G(y^-(\otimes), y_i(\otimes)) = \sum_{j=1}^{m} \gamma_j r_{ij}^- \quad (i = \overline{1,n}).$$

In order to find the weights $\gamma_j, j = \overline{1,m}$, we construct the following bi-objective optimization problem.

$$\max G(y^+(\otimes), y_i(\otimes)) = \sum_{j=1}^{m} \gamma_j r_{ij}^+ \quad (i = \overline{1,n}), \quad \min G(y^-(\otimes), y_i(\otimes)) = \sum_{j=1}^{m} \gamma_j r_{ij}^-,$$

$$s.t. \sum_{j=1}^{m} \gamma_j = 1, \gamma_j \geq 0 \quad (j = \overline{1,m})$$

The problem [P3] can be solved by the optimization problem such as

$$\max D = \sum_{i=1}^{n} \sum_{j=1}^{m} [G(y^+(\otimes), y_i(\otimes)) - G(y^-(\otimes), y_i(\otimes))]\gamma_j,$$

$$s.t. \sum_{j=1}^{m} \gamma_j = 1, \gamma_j \geq 0 \quad (j = \overline{1,m})$$

The final ranking is based on the weighted Borda method by using the rank vectors obtained from the above three methods.

## 5. An illustrative example

Assume that an enterprise manufactures certain equipments and that 5 types of equipment ($A_1$, $A_2$, $A_3$, $A_4$, $A_5$, $A_6$) should be manufactured in the deliberation. The essential attribute which we are going to evaluate is six kinds: the stability ($G_1$), the operation performance ($G_2$), the structure performance ($G_3$), the reliability ($G_4$), the economic worth ($G_5$) and the beautiful view ($G_6$). All of the above six attributes are the effect attribute. Therefore, these attribute values are all the scored values (see Table 1). Their range is from 1 (worst) to 10 (best).

**Table 1**. Decision matrix

|       | $G_1$ | $G_2$ | $G_3$ | $G_4$ | $G_5$ | $G_6$ |
|-------|-------|-------|-------|-------|-------|-------|
| $A_1$ | [6,8] | [8,9] | [7,8] | [5,6] | [6,7] | [8,9] |
| $A_2$ | [7,9] | [5,7] | [6,7] | [7,8] | [6,8] | [7,9] |
| $A_3$ | [5,7] | [6,8] | [7,9] | [6,7] | [7,8] | [8,9] |
| $A_4$ | [6,7] | [7,8] | [6,9] | [5,6] | [8,9] | [7,8] |
| $A_5$ | [7,8] | [6,7] | [6,8] | [5,6] | [5,7] | [7,10] |

By using the algorithm presented in this paper, we find the optimal rank for five kinds of equipment.

Assume that there are two decision-makers and that judgment matrices given by them are as follows.

$$A^{(1)} = \begin{bmatrix} 1 & 2 & 3 & 4 & 5 & 6 \\ 1/2 & 1 & 2 & 3 & 2 & 3 \\ 1/3 & 1/2 & 1 & 2 & 3 & 4 \\ 1/4 & 1/3 & 1/2 & 1 & 2 & 3 \\ 1/5 & 1/2 & 1/3 & 1/2 & 1 & 2 \\ 1/6 & 1/3 & 1/4 & 1/3 & 1/2 & 1 \end{bmatrix}, \quad A^{(2)} = \begin{bmatrix} 1 & 2 & 2 & 4 & 3 & 2 \\ 1/2 & 1 & 2 & 3 & 2 & 3 \\ 1/2 & 1/2 & 1 & 2 & 3 & 4 \\ 1/4 & 1/3 & 1/2 & 1 & 2 & 3 \\ 1/3 & 1/2 & 1/3 & 1/2 & 1 & 2 \\ 1/2 & 1/3 & 1/4 & 1/3 & 1/2 & 1 \end{bmatrix}$$

The subjective preference is given by the grey interval number

$$q(\otimes) = ([1, 3], [2, 3], [1.5, 2], [2.3, 3], [3, 4])$$

The relative approach degree of grey TOPSIS is $C$ = (0.4449, 0.5078, 0.0842, 0.5205, 1.0000). Thus, the rank of plans is such as $A_5 \succ A_4 \succ A_2 \succ A_1 \succ A_3$.

Next, we evaluate the plans by the relative approach degree of grey incidence degree method. For $\theta_+ = \theta_- = 0.5$, $C' = (0.4809, 0.4963, 0.3714, 0.5006, 0.6411)$. Thus, we obtain the rank such as $A_5 \succ A_4 \succ A_2 \succ A_1 \succ A_3$.

Then, we evaluate the plans by the relative membership degree of grey incidence degree method. The membership degree to the ideal plan is $u$ = (0.4618, 0.4926, 0.2588, 0.5013, 0.7614). Thus, the rank of plans is such as $A_5 \succ A_4 \succ A_2 \succ A_1 \succ A_3$.

The final rank by the weighted fuzzy Borda method is $A_5 \succ A_4 \succ A_2 \succ A_1 \succ A_3$.

**Conclusion**

In this paper, for MADM in which all of attribute weights and attribute values are given by interval grey number, we have proposed a new interval weight determining method and three methods of grey interval relation decision-making: the evaluation of plans by the relative approach degree of grey TOPSIS method, the evaluation by the relative approach degree of grey incidence degree method and the evaluation by the relative membership degree of grey incidence degree method. The final rank of plans has been obtained by weighted Borda method considering the above three ranking results.

The features of our method different from other grey relation decision-making methods are as follows. The first feature is finding of the subjective grey interval weights by group AHP, finding of the objective grey interval weights by optimization and entropy method, and then finding of the final grey interval weights by multiplicative composition using the grey interval subjective and objective weights. The second is to obtain the weighted grey interval decision matrix considering the comprehensive grey interval weights determined in the preceding steps for MADM with interval decision matrix. The third is that decision-making is carried out based on the relative approach degree of grey incidence, the relative approach degree of grey TOPSIS and the relative membership degree of grey incidence using 2-dimensional Euclidean distance. The weighted Borda method is used for combining the results of three methods. An example shows the applicability of the proposed approach.